\documentclass[conference]{IEEEtran}
\IEEEoverridecommandlockouts
\usepackage{cite}
\usepackage{amsmath,amssymb,amsfonts}
\usepackage{algorithmic}
\usepackage{graphicx}
\usepackage{textcomp}
\usepackage{xcolor}
\usepackage{fancyhdr}
\def\BibTeX{{\rm B\kern-.05em{\sc i\kern-.025em b}\kern-.08em
    T\kern-.1667em\lower.7ex\hbox{E}\kern-.125emX}}
\begin{document}

\title{Adaptive Per-Tree Canopy Volume Estimation Using Mobile LiDAR in Structured and Unstructured Orchards\\
}

\author{
\IEEEauthorblockN{1\textsuperscript{st} Ali Abedi}
\IEEEauthorblockA{
\textit{Department of Mechanical Engineering} \\
\textit{University of California, Merced} \\
California, United States \\
aliabedi@ucmerced.edu}
\and
\IEEEauthorblockN{2\textsuperscript{nd} Fernando Cladera}
\IEEEauthorblockA{
\textit{Department of Computer and} \\\textit{Information Science} \\
\textit{University of Pennsylvania} \\
Pennsylvania, United States \\
fclad@seas.upenn.edu}
\and
\IEEEauthorblockN{3\textsuperscript{rd} Mohsen Farajijalal}
\IEEEauthorblockA{
\textit{Department of Mechanical Engineering} \\
\textit{University of California, Merced} \\
California, United States \\
mfarajijalal@ucmerced.edu}
\and 
\IEEEauthorblockN{4\textsuperscript{th} Reza Ehsani}\hspace{18.5cm}
\IEEEauthorblockA{
\textit{Department of Mechanical Engineering} \\
\textit{University of California, Merced} \\
California, United States \\
rehsani@ucmerced.edu}
}

\maketitle
\fancypagestyle{withfooter}{
  \renewcommand{\headrulewidth}{0pt}
  \fancyfoot[C]{\footnotesize Accepted to the Novel Approaches for Precision Agriculture and Forestry with Autonomous Robots IEEE ICRA Workshop - 2025}
}
\thispagestyle{withfooter}
\pagestyle{withfooter}

\begin{abstract}
We present a real-time system for per-tree canopy volume estimation using mobile LiDAR data collected during routine robotic navigation. Unlike prior approaches that rely on static scans or assume uniform orchard structures, our method adapts to varying field geometries via an integrated pipeline of LiDAR-inertial odometry, adaptive segmentation, and geometric reconstruction. We evaluate the system across two commercial orchards, one pistachio orchard with regular spacing and one almond orchard with dense, overlapping crowns. A hybrid clustering strategy combining DBSCAN and spectral clustering enables robust per-tree segmentation, achieving 93\% success in pistachio and 80\% in almond, with strong agreement to drone-derived canopy volume estimates. This work advances scalable, non-intrusive tree monitoring for structurally diverse orchard environments.
\end{abstract}

\renewcommand{\thefootnote}{}
\footnotetext{This work was supported primarily by the Internet of Things for Precision Agriculture (IoT4Ag) Engineering Research Center program of the National Science Foundation, USA under NSF Cooperative Agreement No. EEC-1941529.}
\renewcommand{\thefootnote}{\arabic{footnote}}

\begin{figure}[t]
\centering
\includegraphics[width=\columnwidth]{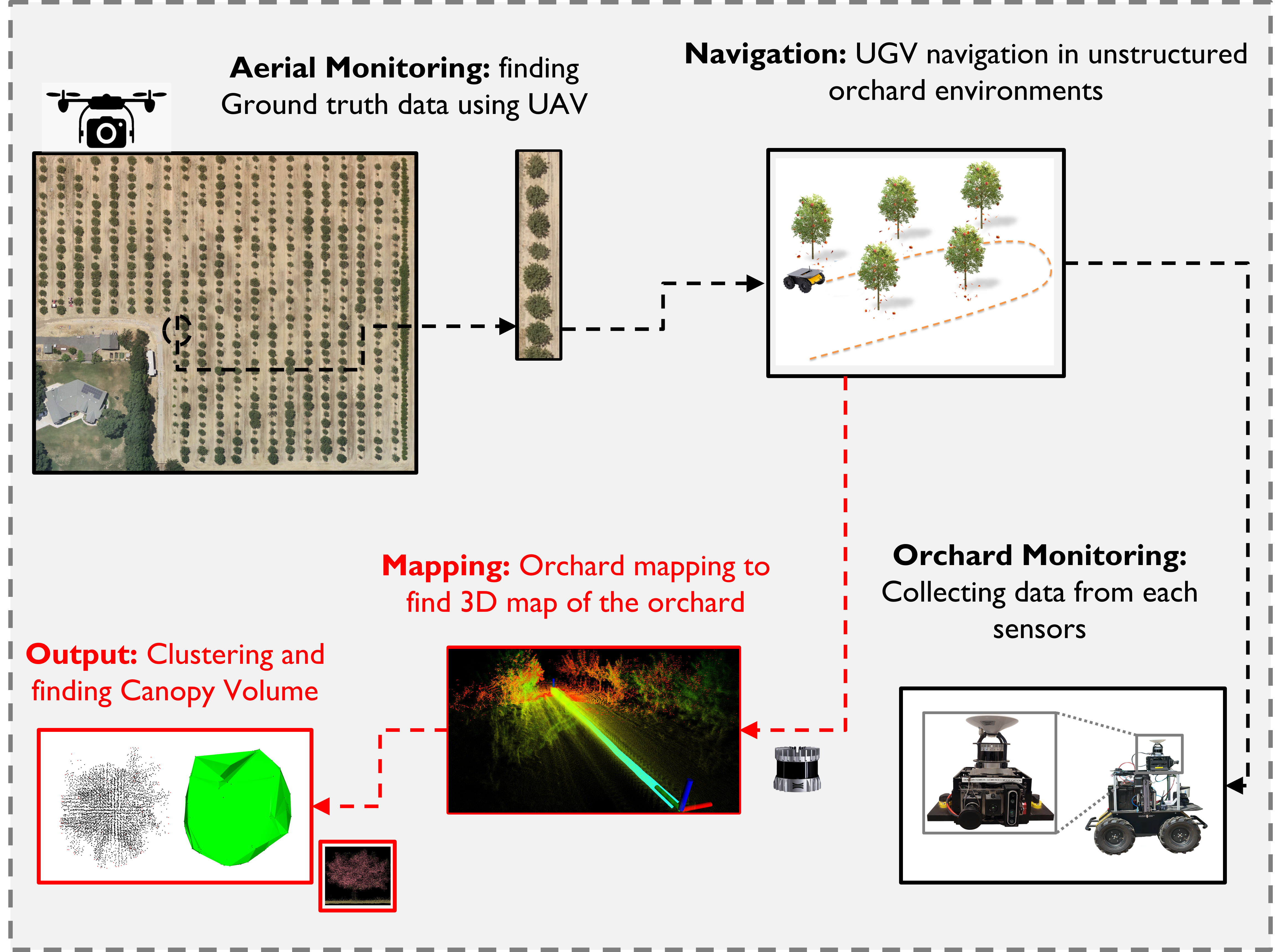}
\caption{Project Overview.}
\label{Overview}
\end{figure}

\section{Introduction}

Accurate estimation of tree canopy volume is fundamental to orchard management, with applications in yield prediction, biomass assessment, and optimized resource allocation~\cite{guo2025study}. Canopy volume serves as a key indicator of tree vigor and productivity, supporting data-driven decisions in precision agriculture~\cite{meng2022health, farhan2024comprehensive}. On the other hand, precise quantification of tree canopy volume and detailed characterization of canopy morphology, particularly for nut species such as almonds and pistachios, play a crucial role in optimizing the efficiency and effectiveness of mechanical harvesting operations \cite{mohsen2025assessing, aragon2023assessment, farajijalal2024data}.

Traditional methods for canopy estimation rely on manual measurements or visual assessments, which are labor-intensive, time-consuming, and prone to human error~\cite{afsah2022mechanical, verma2016comparison}. These limitations hinder their scalability, especially in large commercial operations, and make it difficult to capture the 3D complexity of individual tree structures. To overcome this challenge, researchers have increasingly adopted remote sensing technologies, most notably LiDAR \cite{rivera2023lidar, debnath2023applications} and UAV-based photogrammetry \cite{risal2024improving, abdulridha2023evaluation, dilmurat2022estimating}, which offer high-resolution 3D mapping capabilities~\cite{mcnicol2021extent, zhang2021mapping, ghanbari2021individual, cladera2025air}, as illustrated in Fig.~\ref{Overview}. Among these, ground-based LiDAR platforms offer a unique advantage: they can collect high-fidelity spatial data during routine field navigation, enabling real-time, non-intrusive canopy monitoring. 

While prior studies have demonstrated the utility of LiDAR for canopy geometry estimation~\cite{lee2009laser, rosell2009obtaining}, most rely on uniform orchard layouts with clearly separated trees. However, orchard geometries vary substantially in practice. For example, pistachio orchards tend to have relatively uniform spacing with well-separated canopies, whereas almond orchards often exhibit denser planting, overlapping crowns, and irregular row spacing.

In order to address the existing issues mentioned, the objective of this study is to propose a robust and automated robotic system. The trials were conducted in two distinct environments: a commercial pistachio orchard and a structurally more challenging almond orchard. In the pistachio case, tree segmentation using density-based clustering (DBSCAN) was sufficient. In contrast, the almond orchard posed significant segmentation challenges due to canopy overlap and inconsistent spacing. To address this, we extended our method by integrating graph-based spectral clustering to refine initial groupings and split large, ambiguous clusters.

\begin{figure}[t]
\centering
\includegraphics[width=\columnwidth]{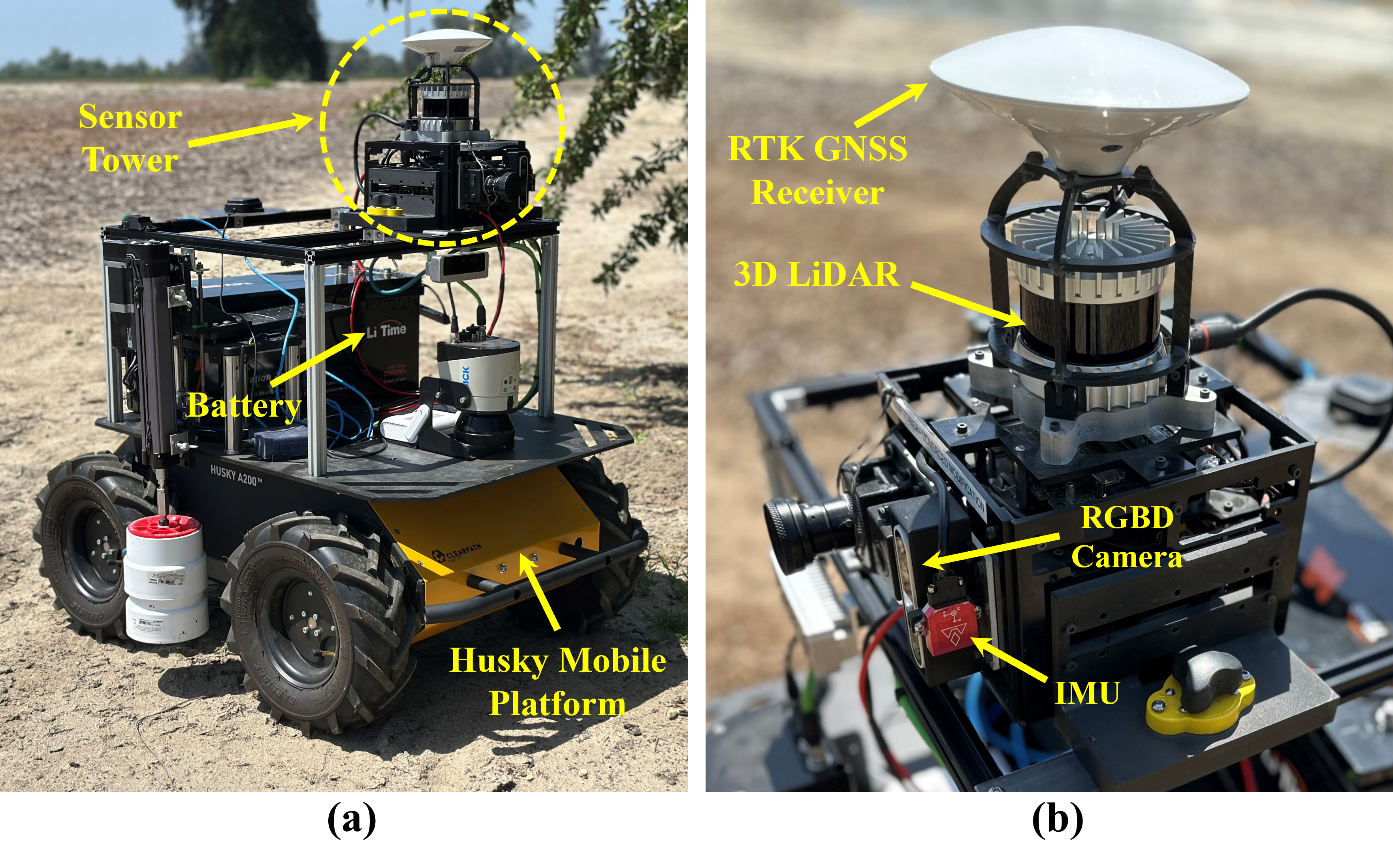}
\caption{The ground mobile LiDAR system setup: 
\textbf{(a)} Sensor tower mounted on the mobile platform; 
\textbf{(b)} Sensor tower system components.
}
\label{fig:system}
\end{figure}

This paper presents a fully automated, real-time canopy volume estimation pipeline capable of adapting to diverse orchard geometries. The proposed mobile LiDAR-based system, illustrated in Fig.~\ref{fig:system}, enables per-tree canopy volume estimation during autonomous navigation through orchard rows, eliminating the need for stationary scanning or offline post-processing. The approach integrates LiDAR-inertial odometry, adaptive clustering techniques, and geometric volume reconstruction using Convex Hull and Alpha Shape algorithms. Experimental results from both orchard environments confirm the system’s robustness and generalizability, highlighting its potential for deployment across varied agricultural landscapes.

\begin{figure*}[t]
\centering
\includegraphics[width=\textwidth]{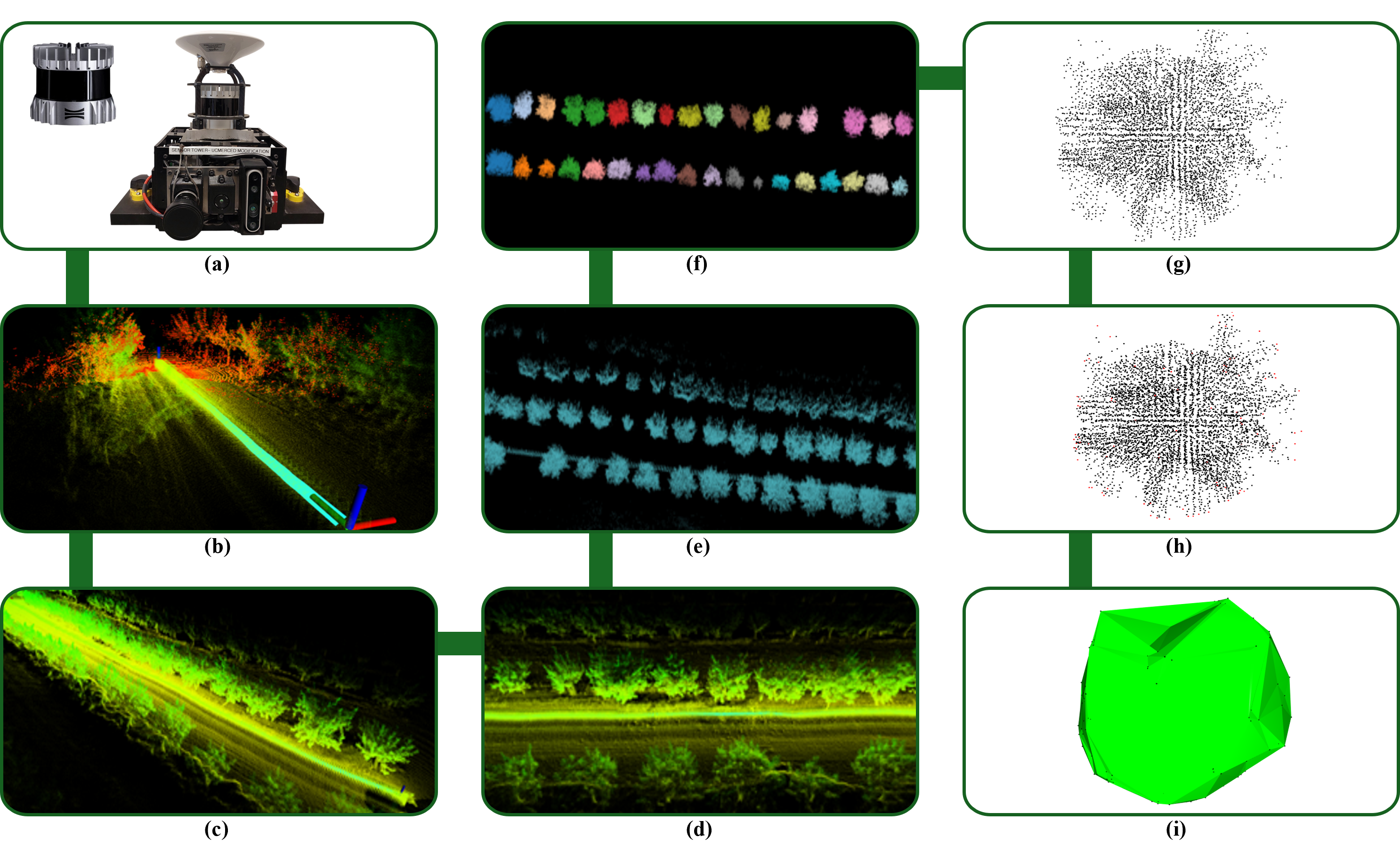}
\caption{Tree canopy volume estimation pipeline: 
\textbf{(a)} Sensor tower mounted on a ground robot collects 3D LiDAR data; 
\textbf{(b)} Raw environment point clouds are read; 
\textbf{(c)} Orchard is mapped (perspective view); 
\textbf{(d)} Orchard is mapped (horizontal view); 
\textbf{(e)} Ground and trunks removed via RANSAC; 
\textbf{(f)} Canopies are segmented, each canopy shown in a different color; 
\textbf{(g)} Voxel-based downsampling is applied to reduce point density; 
\textbf{(h)} Outer boundaries of the canopy are extracted; 
\textbf{(i)} The boundary surface is reconstructed to compute the final volume.}

\label{Total}
\end{figure*}

\section{Materials and Methods}
\subsection{Mobile Sensing Platform and Data Collection}

We used a ground-based mobile platform equipped with a multi-sensor tower to collect spatial data in two commercial orchards: one pistachio orchard and one almond orchard. The sensor suite included a 3D LiDAR, IMU, GPS \cite{maharlooei2024assessing}, and RGB-D cameras. All sensors were synchronized and recorded using an onboard computer running the Robot Operating System (ROS). The robot navigated autonomously through orchard rows during the late growing season, ensuring well-developed canopies at the time of data acquisition.

Although the same sensing configuration was used in both orchards, the environments presented different structural properties. The pistachio orchard featured uniform row spacing and minimal canopy overlap, while the almond orchard exhibited denser planting, frequent crown overlaps, and irregular tree spacing, posing greater segmentation challenges.

\subsection{Pipeline Overview}

The overall canopy volume estimation pipeline consists of four main stages: (1) LiDAR-inertial mapping of the orchard environment; (2) preprocessing and filtering to isolate canopy structures; (3) per-tree segmentation using orchard-specific clustering strategies; and (4) geometric volume estimation using Convex Hull and Alpha Shape reconstructions. The process of the tree canopy volume estimation pipeline is shown in Fig.~\ref{Total}.

The main difference between the two orchard pipelines lies in the canopy segmentation stage. In the pistachio orchard, we used DBSCAN directly. In the almond orchard, overlapping canopies often led to over-clustering, so we introduced a refinement step using spectral clustering to split large ambiguous clusters.

\subsection{Canopy Segmentation Strategies}

\subsubsection{Pistachio Orchard: DBSCAN-Based Segmentation}

For the pistachio orchard, tree canopies were segmented using the DBSCAN algorithm due to its ability to handle non-uniform point distributions. Prior to clustering, point clouds were filtered to remove ground and trunk points using RANSAC-based plane segmentation, followed by voxel downsampling at a resolution of 0.1\,m. DBSCAN was applied with parameters $\varepsilon = 0.8\,\text{m}$ and $N_{\min} = 1300$. This configuration yielded reliable segmentation results, correctly isolating 93\% of tree canopies. Fig.~\ref{Total}f shows an example of successful segmentation using this method.

\subsubsection{Almond Orchard: DBSCAN + Spectral Clustering}

In the almond orchard, the same DBSCAN parameters often resulted in under-segmentation due to canopy overlap and irregular spacing. To address this, we introduced a stricter segmentation strategy. After the initial DBSCAN step, clusters exceeding 45,000 points were identified as likely containing multiple overlapping trees.

These large clusters were further subdivided using spectral clustering based on $k$-nearest neighbor connectivity ($k=10$). The number of subclusters was automatically estimated by dividing the original cluster size by the maximum allowed size. A graph-based spectral embedding was computed, and clustering was performed via $k$-means on the embedded space. This two-stage process, density clustering followed by graph refinement, enabled robust segmentation in highly structured or overlapping canopy environments.

\subsection{Tree Organization and Labeling}

Once segmentation was complete, the centroid of each tree cluster was computed. Trees were grouped into rows based on $y$-axis alignment using linear regression and a distance threshold. Within each row, trees were ordered and labeled sequentially (e.g., $L_0$, $L_1$, $\dots$ or $R_0$, $R_1$, $\dots$), enabling consistent tracking across sessions and environments.

\subsection{Canopy Volume Estimation}

We estimated canopy volume using two geometric techniques: Convex Hull and Alpha Shape. The Convex Hull encloses each point cluster in the smallest convex polyhedron, providing an upper-bound estimate. The Alpha Shape method captures concave regions more accurately by controlling surface tightness with a radius parameter $\alpha$. We used $\alpha = 0.9\,\text{m}$ to balance mesh fidelity and stability. These methods were applied identically in both orchards to ensure comparability across environments.

\subsection{Traversal Speed}
The mobile platform maintained a steady operational speed during data collection, increasing gradually to around $1.5\,\text{m/s}$. Depending on the number of trees per row, traversal time varied from approximately 50 seconds for short rows to around 110 seconds for longer rows. These speeds reflect typical operational conditions in orchard environments and confirm that the system can perform canopy scanning without disrupting normal field activities.

\begin{table*}[htbp]
\caption{Volumes Calculated for Different Trees Using Two Methods and Their Comparison}
\begin{center}
\begin{tabular}{|c|c|c|c|c|c|}
\hline
\textbf{Tree Number} & \textbf{Ground Truth Volume (m$^3$)} & \textbf{Convex Hull (m$^3$)} & \textbf{Convex Hull Error (\%)} & \textbf{Alpha Shape (m$^3$)} & \textbf{Alpha Shape Error (\%)} \\
\hline
1  & 28.06 & 33.26 & 18.55 & 31.49 & 12.24 \\
2  & 23.03 & 26.07 & 13.17 & 24.90 & 8.09 \\
3  & 18.82 & 21.77 & 15.69 & 20.54 & 9.19 \\
4  & 31.30 & 30.23 & 3.43  & 28.75 & 8.15 \\
5  & 27.83 & 27.77 & 0.21  & 26.13 & 6.14 \\
6  & 27.83 & 28.97 & 4.07  & 27.14 & 2.49 \\
7  & 31.30 & 37.59 & 20.11 & 35.10 & 12.14 \\
8  & 12.12 & 15.57 & 14.90 & 28.45 & 22.91 \\
9  & 31.78 & 33.83 & 31.87 & 6.44  & 0.29  \\
10 & 23.23 & 27.07 & 25.74 & 16.54 & 10.80 \\
11 & 25.67 & 24.82 & 23.02 & 3.30  & 10.33 \\
12 & 22.07 & 21.36 & 20.10 & 3.19  & 8.89  \\
13 & 9.53  & 11.38 & 10.89 & 19.48 & 14.35 \\
14 & 26.31 & 26.07 & 24.66 & 0.91  & 6.24  \\
15 & 31.54 & 36.66 & 34.70 & 16.23 & 10.03 \\
16 & 25.04 & 27.08 & 25.32 & 8.11  & 1.08  \\
17 & 18.65 & 19.34 & 18.31 & 3.75  & 1.78  \\
\hline
\end{tabular}
\label{tab:volume_comparison}
\end{center}
\end{table*}

\section{Results}

This section presents the results of our LiDAR-based canopy volume estimation pipeline across two commercial orchards: one with uniformly spaced pistachio trees, and one with denser, irregularly spaced almond trees. While both datasets were acquired using the same mobile sensing platform, the orchard structures presented different segmentation challenges, which we addressed through orchard-specific clustering strategies.

\subsection{Pistachio Orchard: Baseline Performance}

In the pistachio orchard, the DBSCAN-based segmentation pipeline successfully clustered 93\% of the recorded tree canopies. For each detected tree, volume was estimated using both Convex Hull and Alpha Shape reconstructions. As expected, Alpha Shape yielded lower errors due to its ability to capture concave canopy regions more accurately.

Ground truth data was obtained using a DJI Phantom 4 Pro drone and processed through GPS-tagged photogrammetry. Tree volumes were approximated using spherical models based on canopy diameter extracted from orthomosaic maps. Although idealized, this approximation proved sufficient for validation in this orchard, where most tree crowns were approximately symmetrical.

Table~\ref{tab:volume_comparison} shows a sample of seventeen representative trees comparing LiDAR-derived volumes against aerial estimates. The Alpha Shape method consistently outperformed Convex Hull, especially in trees with more complex geometries.

\subsection{Almond Orchard: Segmentation Improvement with Enhanced Clustering}

The almond orchard posed significantly greater segmentation challenges due to overlapping canopies and irregular tree spacing. When the original DBSCAN configuration (used in pistachio) was applied directly, only 46\% of tree crowns were correctly segmented. These failures were primarily due to DBSCAN grouping multiple adjacent trees into a single cluster when their crowns were heavily interwoven.

To address this, we applied the stricter clustering pipeline described in Section II, combining DBSCAN with spectral clustering to automatically subdivide overly large clusters based on spatial connectivity. With this enhanced method, segmentation performance increased substantially, successfully isolating 80\% of individual tree canopies.

Due to the complexity of the almond dataset and partial occlusion in ground truth imagery, we limited quantitative volume analysis to the pistachio dataset. Future work will focus on integrating ground truth annotation pipelines for denser orchard layouts.

\section{Discussion}

The results across two structurally different orchards demonstrate the adaptability and effectiveness of our LiDAR-based system for real-time canopy volume estimation during routine robot navigation. By leveraging data already collected for localization and mapping, the system supports scalable, non-intrusive monitoring without requiring stationary scans or dedicated post-processing.

In the pistachio orchard, where tree canopies are well-separated and planting geometry is relatively uniform, density-based clustering (DBSCAN) achieved high segmentation success. Combined with Alpha Shape reconstruction, this setup produced accurate per-tree volume estimates, closely aligned with aerial ground truth.

However, this success did not generalize to the almond orchard, where canopies frequently overlapped and inter-tree spacing varied substantially. Applying the same DBSCAN parameters in this environment led to over-grouping, where entire rows or sub-rows were merged into single clusters. To address this limitation, we introduced a stricter segmentation approach that combined DBSCAN with graph-based spectral clustering to split overly large clusters based on local connectivity. This adaptive strategy improved canopy segmentation performance from 46\% to 80\%, enabling the system to better handle dense, irregular orchard layouts.

These results highlight a broader insight: \textbf{no single clustering strategy is sufficient across all orchard geometries}. The segmentation method must account for the structural characteristics of the environment, whether that means well-separated trees or overlapping, bushy canopies. Our hybrid approach represents an important step toward generalizable orchard monitoring systems that do not require manual re-tuning for each deployment.

Although segmentation was the main bottleneck in the almond orchard, other challenges remain. Canopy occlusion, partial scans, and underrepresented tree crowns affected both data sets to varying degrees. These limitations suggest future work should explore:
\begin{itemize}
    \item Learning-based clustering or instance segmentation methods that adapt to scene complexity;
    \item Integration of multi-sensor data (e.g., LiDAR + RGB or thermal) for improved disambiguation in dense environments;
    \item Automated parameter tuning using point density or canopy shape priors.
\end{itemize}

Finally, organizing the segmented trees into a consistent spatial framework supports longitudinal studies. This opens the door to tracking growth patterns over time, detecting structural changes, and developing predictive models for yield or health monitoring, regardless of orchard layout.

Overall, this work demonstrates that accurate and scalable canopy volume estimation is achievable in diverse orchard conditions when the segmentation pipeline is tuned to the geometry of the environment. Our system generalizes beyond a single orchard type, making it a strong candidate for deployment in real-world agricultural monitoring at scale.

\section{Conclusion}

We presented a real-time LiDAR-based system for the estimation of the canopy volume per tree, tested in two orchards with contrasting structures. In the pistachio orchard, DBSCAN-based clustering enabled high segmentation accuracy and reliable volume estimates. In the more complex almond orchard, we introduced a hybrid method combining DBSCAN with spectral clustering, which improved the segmentation performance from 46\% to 80\%.

These results show that adaptable segmentation is essential for generalizing canopy monitoring across diverse orchard geometries. The proposed system offers a scalable, non-intrusive solution for automated, per-tree structural analysis in precision agriculture. Future work will explore multisensor fusion and long-term canopy tracking.

\bibliographystyle{IEEEtran}
\bibliography{References}

\end{document}